\documentclass[conference]{IEEEtran}
\IEEEoverridecommandlockouts
\usepackage{cite}
\usepackage{amsmath,amssymb,amsfonts}
\usepackage{tabularx}
\usepackage{algorithmic}
\usepackage{graphicx}
\usepackage{textcomp}
\usepackage{xcolor}
\usepackage{siunitx}
\usepackage{nicefrac}
\usepackage{multirow}
\usepackage[T2A,T1]{fontenc}
\usepackage[utf8]{inputenc}
\usepackage[russian,english]{babel}
\usepackage[hidelinks]{hyperref}
\newcommand{\atosensedata}{ATOSenseData}
\def\BibTeX{{\rm B\kern-.05em{\sc i\kern-.025em b}\kern-.08em
    T\kern-.1667em\lower.7ex\hbox{E}\kern-.125emX}}
\begin{document}

\title{Measuring Train Driver Performance as Key to Approval of Driverless Trains\\
\thanks{This work is funded by the DZSF. This is not an official statement, legal advice, guideline or directive of the German Federal Railway Authority. \textcopyright 2025 IEEE. Personal use of this material is permitted. Permission from IEEE must be obtained for all other uses, in any current or future media, including reprinting/republishing this material for advertising or promotional purposes, creating new collective works, for resale or redistribution to servers or lists, or reuse of any copyrighted component of this work in other works. DOI: \href{https://ieeexplore.ieee.org/document/11219606}{10.1109/IAVVC61942.2025.11219606}}
}

\author{\IEEEauthorblockN{1\textsuperscript{st} Rustam Tagiew}
\IEEEauthorblockA{\textit{German Centre for Rail Traffic Research (DZSF)} \\
\textit{at the Federal Railway Authority}\\
Dresden, Germany\\
TagiewR@dzsf.bund.de}
\and
\IEEEauthorblockN{2\textsuperscript{nd} Prasannavenkatesh Balaji}
\IEEEauthorblockA{\textit{German Centre for Rail Traffic Research (DZSF)} \\
\textit{at the Federal Railway Authority}\\
Dresden, Germany \\
BalajiP@dzsf.bund.de}
}

\maketitle

\begin{abstract}
Points 2.1.4(b), 2.4.2(b) and 2.4.3(b) in Annex I of Implementing Regulation (EU) No. 402/2013 allow a simplified approach for the safety approval of computer vision systems for driverless trains, if they have 'similar' functions and interfaces to the replaced human driver. The human driver is not replaced one-to-one by a technical system - only a limited set of cognitive functions are replaced. However, performance in one of the most challenging tasks, obstacle detection, is difficult to quantify due to the deficiency of published measurement results. This article summarizes the data published so far and goes a long way to remedy this situation by providing a new public and anonymised dataset of 711 train driver performance measurements from controlled experiments. The measurements are made for different speeds, obstacle sizes, train protection systems and obstacle colour contrasts respectively. The measured values are reaction time and distance to the obstacle. The goal of this paper is an unbiased and exhaustive description of the presented dataset for research, standardization and regulation. The dataset with supplementing information and literature is published on FID move (https://data.fid-move.de/dataset/atosensedata).

\end{abstract}
\begin{IEEEkeywords}
obstacle detection, driverless train operation, reaction time, train protection system.
\end{IEEEkeywords}

\section{Introduction}
Driverless train operation (DTO) and unattended train operation (UTO) on open tracks is a major international innovation domain~\cite[p. 96]{bmdv}. It offers several benefits to the railway industry. Therefore, it is important for authorities involved to remove obstacles in its development. Replacing a human driver with a technical system on open tracks requires advanced computer vision (CV) systems for obstacle detection and other functions. This is different from driverless trains on isolated tracks, such as metros, where obstacles are eliminated by physical barriers. The safety approval of CV systems for mainline railways (in the European context) for DTO and UTO is a critical step towards enabling their regular operation.

A certifying authority would need performance targets against which a CV system should be tested. Human performance is not only intuitively, but also legally, as Section~\ref{legal} will show, a legitimate source for such performance targets. In order to set performance targets based on human performance, it is necessary to measure it. The DZSF-funded project ``Functional requirements for sensors and logic of an ATO unit'' (ATO-Sense)~\cite{ato-sense} conducted simulator studies with human subjects. The resulting dataset \atosensedata~\cite{atosensefimove} and its description in the context of the approval of driverless trains is the main contribution of this paper. 

The methodology used in ATO-Sense is described in Section~\ref{ato-sense}. Related work on human performance measurement is summarised in Section~\ref{related}. Section~\ref{stats} presents observational and non-parametric statistics on the dataset generated within ATO-Sense, \atosensedata. Finally, Section~\ref{discussion} explains the value of the contribution provided by \atosensedata~as a key to the approval of driverless trains.

\section{Summary Of The Legal Environment}
\label{legal}
\subsection{CSM-RA is Central to the Approval}
For European mainline railways, the approval process for DTO and UTO has to be built on Art. 21 of the Interoperability Directive (EU) 2016/797 and the Implementing Regulation on Approval (EU) 2018/545 issued to refine it~\cite{tagiew2021towards,spec99004,spec91516}. According to Art. 21(3) of the Interoperability Directive, a vehicle has to satisfy four requirements:
\begin{itemize}
\item[a)] Subsystems meet the essential requirements set by Interoperability Directive (EU) 2016/797, Annex III.
\item[b)] Subsystems are compatible with each other.
\item[c)] Subsystems are safely integrated into the vehicle.
\item[d)] The vehicle is compatible with the network.
\end{itemize}

These requirements are established by the relevant Technical Specifications for Interoperability (TSIs) and notified national rules. A CV system would be categorized as the ``on-board control-command and signalling''~\cite{tagiew2021towards,spec99004,spec91516}, which is defined in the Interoperability Directive (EU) 2016/797, Annex II, Section 2.4. It describes ``all the on-board equipment required to ensure safety and to command and control movements of trains authorized to travel on the network''. According to Implementing Regulation (EU) 2023/1695, Control Command and Signalling (CCS) TSI would be applicable. CCS TSI, 2.2 in Annex 1 requires that all CCS systems are assessed in accordance with the Commission Implementing Regulation (EU) 402/2013, i.e. the Common Safety Methods for the Evaluation and Assessment of Risks (CSM-RA).\\

\subsection{Human As A Reference System According To CSM-RA}
CSM-RA Art. 3(20) defines a reference system as:
\begin{quote}
``a system proven in use to have an acceptable safety level and against which the acceptability of the risks from a system under assessment can be evaluated by comparison''
\end{quote}
A human train driver can be considered as a reference system~\cite{tagiew2021towards,atorisk}. According to CSM-RA Annex I point 2.4.1, a similar system can be considered as a reference system, if ``one, several or all hazards are appropriately covered by it''. Therefore, the comparison with a reference system is the approval principle (b) as according to point 2.1.4 that states:
\begin{quote}
``The risk acceptability of the system under assessment shall be evaluated by using one or more of the following risk acceptance principles:
\begin{itemize}
\item[(a)] the application of codes of practice (point 2.3);
\item[(b)] a comparison with similar systems (point 2.4);
\item[(c)] an explicit risk estimation (point 2.5);''
\end{itemize}
\end{quote}

A human driver will probably not be replaced by a AI-based robot for a driverless train. Nor does it seem necessary to demand a one-to-one mapping of the rather complex cognitive skills of the train driver by a CV system~\cite[pp. 9,10]{polz}. Fortunately, CSM-RA Annex I point 2.4.2 demands just similar and not equal functions and interfaces as follows:

\begin{quote}
``A reference system shall satisfy at least the following requirements:
\begin{itemize}
\item[(a)] it has already been proven in use to have an acceptable safety level and would therefore still qualify for approval in the Member State where the change is to be introduced;
\item[(b)] it has similar functions and interfaces as the system under assessment
\item[(c)] it is used under similar operational conditions as the system under assessment;
\item[(d)] it is used under similar environmental conditions as the system under assessment.''
\end{itemize}
\end{quote}

To ensure similar functions, it is necessary to compile a list of the functions of a human driver and their implementation in a technical system~\cite{spec91516}. While functional output interfaces are obviously similar~\cite{spec91516}, the issue is more complicated for functional input interfaces. There is a consensus that RGB cameras and the human eye are considered similar~\cite{spec91516}. CV systems with RGB cameras as their sensors therefore need a rather simple performance comparison with the human driver for each implemented function. According to CSM-RA Annex I point 2.4.3, the performance of the human driver may be derived from ``the safety analyses or from an evaluation of safety records''.    

According to point 2.4.4, if the functional input interface is not similar, i.e. the CV system is different, at least the same level of safety as the reference system shall be achieved by ``applying another reference system or one of the two other risk acceptance principles''. Again, this emphasizes that human performance should be measured.

\section{Related Work}
\label{related}
For a legally sound approval, the multitude of human abilities involved in the task of monitoring the track must be distributed into functions. The current consensus~\cite{spec91516} suggests that the functions of object detection and obstacle classification are central to this task. One metric that is well established in railway standards, in the context of human performance is the human error probability (HEP)~\cite{vde103}. HEP is the error rate of a human operator performing a task. There is a subtle difference between a task and a function. Tasks have steps and goals whereas functions have inputs and outputs. In order to apply HEP to a function, it must correspond to a task. Although HEP is applied in the same way to missed obstacles and signals passed at risk, several works propose non-binary, i.e. continuous, metrics for obstacle detection~\cite{tagiew2023mainline}. 

Three such continuous measures are the distance to the obstacle at the time of its detection, the time to arrive at the time of its detection, and the reaction time (RT), i.e., the difference between the time of the obstacle's appearance and the time of its detection. While they all have their advantages and disadvantages, distance to the obstacle is not applicable for suddenly appearing obstacles. Time to arrival is highly dependent on speed, and for RT, the time of obstacle appearance is difficult to determine. The choice of metric is orthogonal to the choice of the measurement method. According to point 2.4.3 of CSM-RA Annex I, ``safety analyses'' and ``evaluation of safety records'' can be used as measurement methods. Evaluation of safety records is the most realistic source of human performance values~\cite{spec91516}. Unfortunately, publicly available records of human error, i.e. incidents and accidents, are rare and not willingly published by railway operators. Simulator studies and field experiments are less realistic, especially considering that obstacles are rarer in reality than in experiments.

Human reliability analysis (HRE) is a theoretical method of determining HEPs described in railway standards~\cite{vde103} and is considered to be the least realistic~\cite{spec91516}. The report of the ATO-Risk project, funded by the DZSF, contains a table of HEPs determined by HRA for different cases of obstacle detection~\cite[pp. 74,75]{atorisk}. Binary measures of human performance can also be obtained from safety records. Statistical analysis of accident data from an undisclosed European region from January 2022 to January 2024 provided such results~\cite{lahneche2024analysing}. According to the study, human drivers can avoid about $\qty{28,43}{\%}$ ($\qty{95}{\%}$ confidence interval between $\qty{15,70}{\%}$ and $\qty{39,80}{\%}$) of collisions in good visibility conditions. Only $\qty{11,81}{\%}$ could be avoided in all conditions. Distance to object or distance to obstacle at detection is a relatively common measure of human driver performance. Table~\ref{distances} presents median or mean distances to obstacle indicating human driver performance. As seen, the measurement methods differ depending on the source and geographical origin, which might be an issue when using them according to point 2.4.2(c,d) of CSM-RA Annex I.
\begin{table}[t]
  \centering
  \caption{Human detection of objects on tracks~\cite{tagiew2023mainline}.}
  \begin{tabular}{|l|c|c|c|}
    \hline
        \multirow{2}{*}{\textbf{Object Size, Conditions, Reference}}
     & \multicolumn{3}{c|}{\textbf{Distance to Object ($\unit\metre$)}}\\\cline{2-4}
    &  \multicolumn{3}{c|}{\textbf{Median}} \\
    \hline
    Rectangle, $\geq\qty{0.4}{\metre^2}$, $\qty{30}{\percent}$ Contrast & \multicolumn{3}{c|}{$>750$ } \\
    Rectangle, $\qty{2}{\metre^2}$, $\qty{8}{\percent}$ Contrast & \multicolumn{3}{c|}{$500$} \\
    Rectangle, $\qty{0.4}{\metre^2}$, $\qty{8}{\percent}$ Contrast & \multicolumn{3}{c|}{$240$} \\
    Rectangle, $\qty{2}{\metre^2}$, $\qty{30}{\percent}$ Contrast, Night  & \multicolumn{3}{c|}{ $180$} \\
    Rectangle, $\qty{0.4}{\metre^2}$, $\qty{30}{\percent}$ Contrast, Night  & \multicolumn{3}{c|}{ $60$ }  \\
    Rectangle, $\leq\qty{2}{\metre^2}$, $\qty{8}{\percent}$ Contrast, Night & \multicolumn{3}{c|}{$<60$} \\
    \cline{2-4}
    \multicolumn{4}{|l|}{Contrasts are visual}\\
    \multicolumn{4}{|l|}{German field experiments~\cite{polz}}\\
    \hline
    Cube, $40\times40\times\qty{40}{\centi\metre}$ & \multicolumn{3}{c|}{$250$ } \\
    Cube, $20\times20\times\qty{20}{\centi\metre}$ & \multicolumn{3}{c|}{$175$} \\
    Cube, $10\times10\times\qty{10}{\centi\metre}$ & \multicolumn{3}{c|}{ $50$ } \\
    Cube, $5\times5\times\qty{5}{\centi\metre}$ & \multicolumn{3}{c|}{$<25$}  \\
    \cline{2-4}
    \multicolumn{4}{|l|}{Fluorescent objects, $\qty{60}{ \nicefrac{\kilo\metre}{\hour}}$, Night}\\
    \multicolumn{4}{|l|}{Japanese field experiments~\cite{itoh}}\\
    \hline
    Person in safety jacket & \multicolumn{3}{c|}{$400$} \\
    Passenger car & \multicolumn{3}{c|}{$300$} \\
    Person & \multicolumn{3}{c|}{$240$} \\
    Passenger car, Night & \multicolumn{3}{c|}{$<60$ }  \\
    Person with safety jacket, Night & \multicolumn{3}{c|}{$<60$}\\ 
    \cline{2-4}
    \multicolumn{4}{|l|}{German field experiments~\cite{mockel2003multi}}\\
    \hline
     
                     & \textbf{Min.} & \textbf{Median} & \textbf{Max.}\\
    \hline 
    Tree, $50$-$\qty{70}{\nicefrac{\kilo\metre}{\hour}}$ & $50$ & $60$ & $130$\\
    Fallen rock, $20$-$\qty{120}{\nicefrac{\kilo\metre}{\hour}}$ & $7$ & $30$ & $100$ \\
    \cline{2-4}
    \multicolumn{4}{|l|}{Japanese accident statistics~\cite{nakasone2017frontal} }\\
    \hline
                         & \textbf{Min.} & \textbf{Mean} & \textbf{Max.} \\
     \hline
     Pedestrian target (adult), $\qty{1.88}{\metre}$ high       &    $147$ & $323$ &  $600$      \\
     Pedestrian target (child), $\qty{1.28}{\metre}$ high       &     $106$ & $186$ & $320$       \\
         \cline{2-4}
      \multicolumn{4}{|l|}{Visibility $>\qty{10000}{\metre}$, Illumination $>\qty{50}{\lux}$}\\
      \multicolumn{4}{|l|}{Russian field experiments~\cite{popov}}\\
     \hline
  \end{tabular}
  \label{distances}
\end{table}

\section{ATO-Sense Dataset}
\label{ato-sense}
\subsection{ATO-Sense Project}
The DZSF is a departmental research facility of the German Federal Government and one of its goals is to prepare the regulatory structure for the approval of driverless trains. It had issued a tender for ATO-Risk (2020-2024), whose goal was to determine an approach for recording the performance of the human senses as a basis for defining the requirements for a technical ATO system~\cite{atosense,eiart}. The project was awarded to a consortium of Bahnbetrieb und Infrastruktur (BBI) at the TU Berlin in cooperation with the German Aerospace Center (DLR), Siemens Mobility and DB Systemtechnik.
\subsection{Methodology}
\begin{figure*}[t]
  \centerline{\includegraphics[width=1\linewidth]{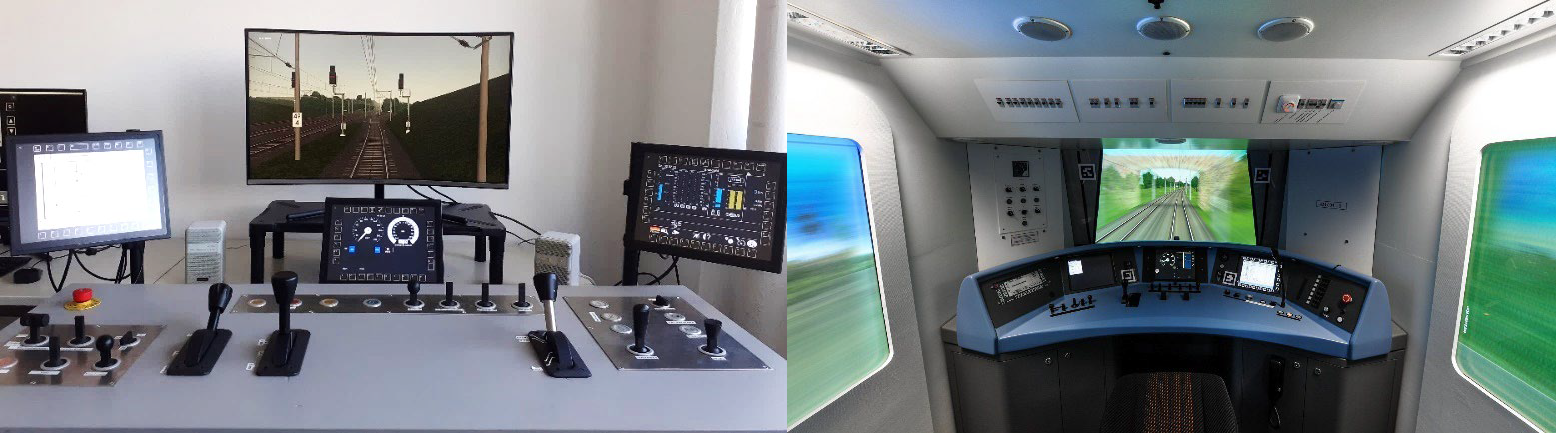}}  
  \caption{Simulation environments of \atosensedata~\cite{atosense,eiart}. On the left side, the simulation environment of BBI resembles a driver's cab of the 193/248 series (Vectron Dual Mode). The software used is Zusi 3 Professional with the display being a 32-inch UHD screen. On the right side, DLR used the simulation environment RailSET{\textregistered}~\cite{johne2016railsite}. An original control panel of a traction unit together with the VIRES software were used. In addition to the front screen, views from the side windows and ambient sounds were provided.}
  \label{simulatoren}
\end{figure*}

Since about $\qty{69}{\%}$ of the tasks of human train drivers are related to visual perception~\cite{atosense,eiart}, the experiments focused on the sense of vision. BBI and DLR have each conducted studies with human subjects in two different train driving simulation environments, as shown in Fig.\ref{simulatoren}. Errors while performing the task of monitoring the tracks have the most dramatic impact of all tasks performed by train drivers. Therefore, stimuli were presented at the tracks in the simulations under different operational and environmental conditions. The stimuli consisted of cubes representing various objects including people. They were always projected as squares to the eyes of the subjects, since they were only presented on straight railway sections with almost no inclination in the conducted experiments.

\begin{figure*}[t]
  \centerline{\includegraphics[width=\linewidth]{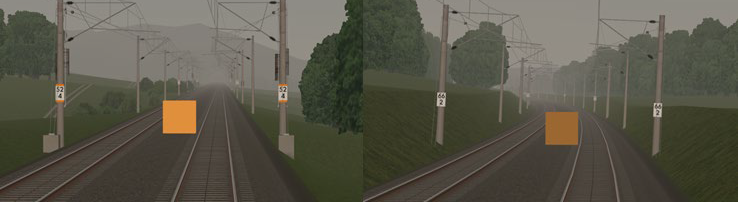}}  
  \caption{Contrast of stimuli in the BBI simulation environment in \atosensedata~\cite{atosense,eiart}, high on the left and low on the right side. Right picture shows a curved track, which did not appear in experiments.}
  \label{contrast}
\end{figure*}

After data cleansing, a total of $424$ recorded human decisions from $25$ DLR participants and $287$ recorded human decisions from $18$ participants in the BBI study were generated. Both studies were conducted on qualified train drivers~\cite{ato-sense}. For BBI, the participating drivers had an average $7.3$ years of professional experience and were $33.4$ years old, on average. For DLR, they had an average of $9.9$ years of professional experience and were 33.7 years old on average. The measurements are made for different speeds ($\qty{40}{\nicefrac{\kilo\metre}{\hour}}$, $\qty{100}{\nicefrac{\kilo\metre}{\hour}}$ and $\qty{160}{\nicefrac{\kilo\metre}{\hour}}$), stimuli sizes ($\qty{90}{\centi\metre}$ and $\qty{180}{\centi\metre}$), train protection systems (FAS, PZB and ETCS) and different obstacle colour contrasts (high and low). These are the performance shaping factors (PSFs) present in the studies. FAS or no train protection stands for ``Fahren auf Sicht'' (Driving By Sight), PZB for Point Train Control and ETCS for European Train Control System. As can be seen in the Fig.\ref{contrast}, high contrast objects had the colour HEX-Code \colorbox[HTML]{F18E2A}{\#f18e2a}, resembling the colour of German railway safety jackets, and the low contrast objects with the colour \colorbox[HTML]{9D6830}{\textcolor{white}{\#9d6830}}.

The participating train drivers were asked to actuate and sound the train horn, once they see an orange or brown cube. It was explained to them that no collision with the cube will occur and they do not have to brake for the cube. In the BBI study, the cubes were placed at a maximum distance of $\qty{3}{\metre}$ to the right or left of the tracks' center~\cite{eiart}. The cubes in the DLR study were placed at the centre of the tracks and moved away from the tracks shortly before collision. Additionally in the BBI study, the rate of false negatives (where the drivers missed to sound the horn before a detectable object was passed by) was found to be $9.24$\% (with the individual rates being $6.5$\%, $6.9$\% and $17.9$\%, for ETCS, PZB and FAS respectively). The PZB and ETCS routes were always completed first and routes with no train protection last. The order of PZB and ETCS routes was equalised between the participants and the stimuli appeared once the train approached them as close as approximately $\qty{800}{\metre}$. Due to the differences between the simulation environments, the visual angle at the first appearance of the stimuli for subjects had a variation (Table~\ref{angles}).

\begin{table}[htbp]
\caption{Visual angle for stimuli }
\begin{center}
\begin{tabular}{|c|c|c|}
\hline
\textbf{Simulator} & \textbf{Size ($\unit\cm)$} & \textbf{Angle ($\unit\min$)} \\
\hline
 \multirow{2}{*}{BBI} & $90$ & $\ang{;3.78;}$ \\
 \cline{2-3}
                      & $180$  & $\ang{;7.56;}$ \\
\hline
\multirow{2}{*}{DLR} & $90$  & $\ang{;6.72;}$ \\
 \cline{2-3}
                      & $180$  & $\ang{;13.43;}$ \\
\hline
\end{tabular}
\label{angles}
\end{center}
\end{table}

The BBI study used an approximately~$\qty{31.5}{\kilo\metre}$ long ($25$-$30$ minutes) route for $8$ stimuli, for PZB, an approximately $\qty{50}{\kilo\metre}$ long ($25$-$30$ minutes) route with $5$ stimuli and for ETCS as well as FAS an approximately $\qty{6}{\kilo\metre}$ long ($10$-$15$ minutes) route with $4$ stimuli. The DLR study used an approximately $\qty{31}{\kilo\metre}$ (approx. $35$ minutes) route for $8$ stimuli, a $\qty{100}{\kilo\metre}$ (approx. $35$ minutes) route with $8$ stimuli and a $\qty{5}{\kilo\metre}$ (approx. $10$ minutes) route with $2$ stimuli for PZB, ETCS and FAS respectively~\cite{atosense}. 

Table~\ref{decisions} shows the number of recorded human decisions distributed over different combinations of PSFs. Speeds and train protection systems are interconnected. No train protection for speeds higher than $\qty{40}{\nicefrac{\kilo\metre}{\hour}}$ and PZB for speeds higher than $\qty{100}{\nicefrac{\kilo\metre}{\hour}}$ are not realistic. On the other side, ETCS is not used for $\qty{40}{\nicefrac{\kilo\metre}{\hour}}$, although this happens in reality. Some predictable PSF configurations are missing from the recordings due to various shortcomings during the fragile experiments. For every recorded human decision in Table~\ref{decisions}, the RT, the distance to the obstacle and the time to arrival are measured. Due to sudden appearance of the stimuli in a predefined distance, RT can be considered to be the most expressive metric for the presented experiments~\cite{spec91516}.

\begin{table}[htbp]
\caption{Number of recorded human decisions depending on PSFs}
\begin{center}
\begin{tabular}{|c|c|c|c|c|c|}
  \hline
  \multirow{2}{*}{\textbf{Speed}} & \textbf{Visual} & \multirow{2}{*}{\textbf{Contrast}} & \multicolumn{3}{|r|}{\textbf{Train Protection System}} \\
\cline{4-6} 
 & \textbf{Angle} &  & \textbf{FAS} & \textbf{PZB} & \textbf{ETCS} \\ 
  \hline
  \multirow{6}{*}{$\qty{40}{\nicefrac{\kilo\metre}{\hour}}$} & \multirow{2}{*}{$\ang{;3.78;}$} & Low &  $17$ & & \\   \cline{3-6}
  &   &   High &  $16$ & & \\    \cline{2-6}
  & \multirow{2}{*}{$\ang{;7.56;}$}   & Low &  $16$ & &\\    \cline{3-6}
  &                           & High &  $15$ & &\\    \cline{2-6}
  & \multirow{2}{*}{$\ang{;13.43;}$}   & Low &  $25$ & &\\    \cline{3-6}
  &                            & High &  $25$ & & \\    \cline{2-6}
  \hline
  \multirow{8}{*}{$\qty{40}{\nicefrac{\kilo\metre}{\hour}}$} & \multirow{2}{*}{$\ang{;3.78;}$} & Low & & $17$ & \\    \cline{3-6}
  &                         &  High & & $17$ & \\    \cline{2-6}
  & \multirow{2}{*}{$\ang{;7.56;}$} &  Low & & $16$ & \\    \cline{3-6}
  &                         &  High & & $15$ & \\    \cline{2-6}
  & \multirow{2}{*}{$\ang{;6.72;}$} &  Low & & $23$ &\\    \cline{3-6}
  &        &  High & & $23$ & \\    \cline{2-6}
  & \multirow{2}{*}{$\ang{;13.43;}$} &  Low & & $23$ & \\    \cline{3-6}
  &                         &  High & & $23$ & \\    \cline{2-6}
  \hline
  \multirow{8}{*}{$\qty{100}{\nicefrac{\kilo\metre}{\hour}}$} & \multirow{2}{*}{$\ang{;3.78;}$} &  Low & & $17$ &\\    \cline{3-6}
  &    &  High & & $18$ &  \\    \cline{2-6}
  & \multirow{2}{*}{$\ang{;7.56;}$} &  Low & & $18$ & \\    \cline{3-6}
  &                         &  High & & $17$ & \\    \cline{2-6}
  & \multirow{2}{*}{$\ang{;6.72;}$} &  Low & & $23$ &\\    \cline{3-6}
  &                         &  High & & $23$ & \\    \cline{2-6}
  & \multirow{2}{*}{$\ang{;13.43;}$} &  Low & & $23$ & \\    \cline{3-6}
  &                          &  High & & $23$ & \\    \cline{2-6}
  \hline
  \multirow{7}{*}{$\qty{100}{\nicefrac{\kilo\metre}{\hour}}$} & \multirow{2}{*}{$\ang{;3.78;}$} &  Low & & & $16$ \\    \cline{3-6}
  &            &  High & & & $18$ \\    \cline{2-6}
  & \multirow{2}{*}{$\ang{;6.72;}$} &  Low & & & $23$ \\    \cline{3-6}
  &                         &   High & & & $23$ \\    \cline{2-6}
  & $\ang{;7.56;}$ &  Low & & & $18$ \\ \cline{2-6}
  & \multirow{2}{*}{$\ang{;13.43;}$} &  Low & & & $24$ \\    \cline{3-6}
  &  &  High & & & $24$ \\    \cline{2-6}
  \hline
  \multirow{6}{*}{$\qty{160}{\nicefrac{\kilo\metre}{\hour}}$} & \multirow{2}{*}{$\ang{;3.78;}$} &  Low & & & $18$ \\    \cline{3-6}
  &  &  High & & & $18$ \\    \cline{2-6}
  & \multirow{2}{*}{$\ang{;6.72;}$} &  Low & & & $24$ \\    \cline{3-6}
  &   &  High & & & $24$ \\    \cline{2-6}
  & \multirow{2}{*}{$\ang{;13.43;}$} &  Low & & & $24$ \\    \cline{3-6}
  &  &  High & & & $24$ \\    \cline{2-6}
   \hline
\end{tabular}
\label{decisions}
\end{center}
\end{table}

\section{Observational Statistics of \atosensedata}
\label{stats}
\begin{figure*}[htbp]
  \centerline{\includegraphics[height=0.92\textheight, width=\linewidth]{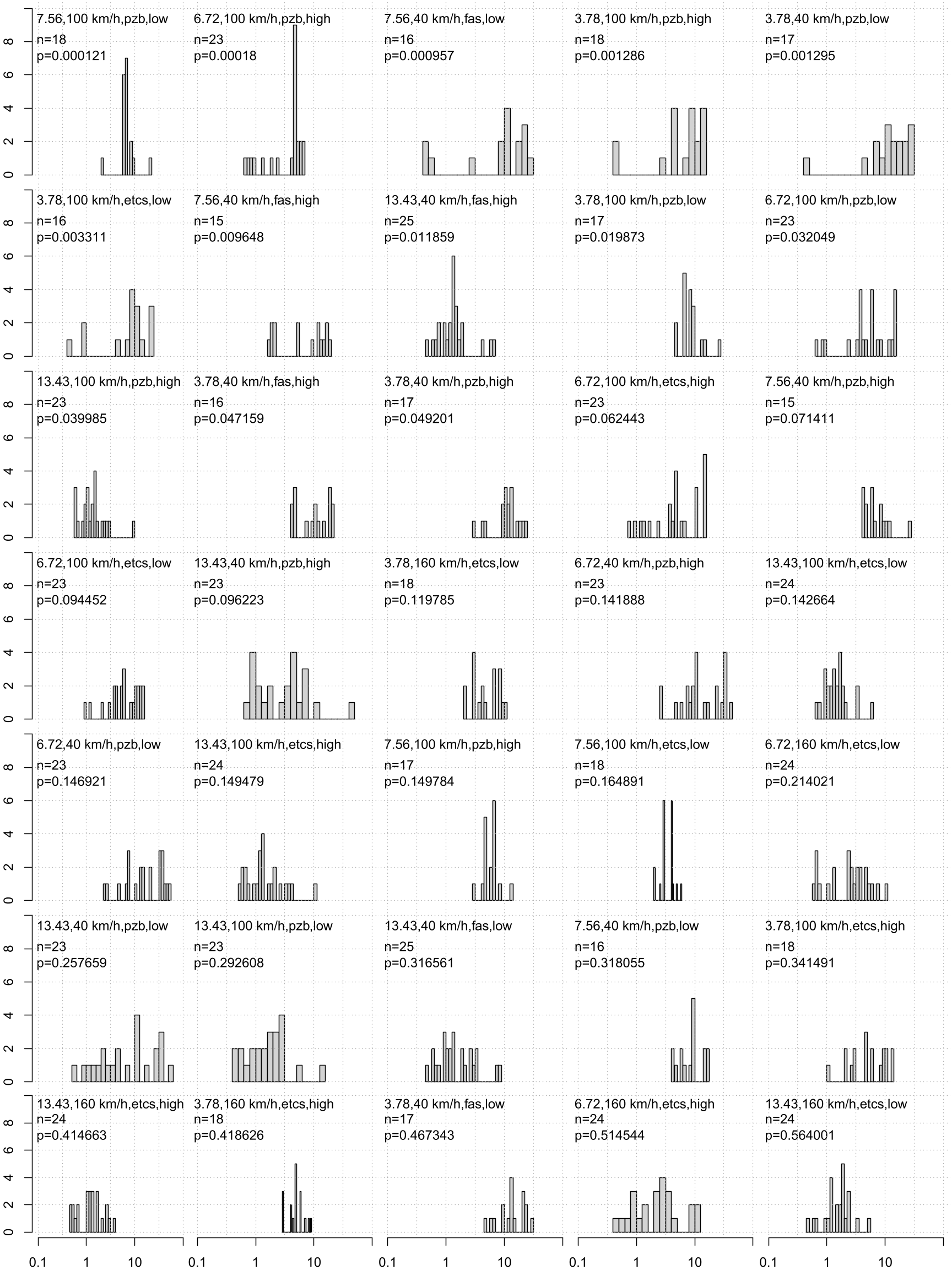}}  
  \caption{Shapiro-Wilk normality test of logarithmic RT for every configuration of four PSFs. The PSFs in a configuration are denoted by a comma-separated list of their values as visual angle, speed, train protection system and contrast. The x-axis plots the logarithmic RT in seconds and y axis the absolute number of samples. $n$ refers to the number of samples and $p$ denotes the p-value for normal distribution.}
  \label{shapirort}
\end{figure*}

Since the goal of this paper is an unbiased and exhaustive description of the presented dataset for research, standardisation and regulation, only observational and non-parametric statistics will be presented here for reference. Due to the non-stationary nature of stimuli, only the RT metric is considered. The underlying behaviour of human perception and reaction to approaching stimuli are not fully understood, making it difficult to determine the true phenomena behind the distribution of RT. Figure~\ref{shapirort} shows the histograms for logarithmic RTs from all $35$ PSF configurations from Table~\ref{decisions}. For the given small samples of PSF configurations, the most powerful normality test, Shapiro-Wilk~\cite{razali2011power}, is used to test the normality of the histograms. The histograms are sorted in ascending order by their p-values. 

The results in Figure~\ref{shapirort} show that the violation of the normality of logarithmic RT cannot be ruled out. The use of such non-parametric and unpaired statistical tests for ordinal data such as the Mann-Whitney $U$ test and the Kruskal-Wallis $H$ test would be recommended to indicate statistical significance for the influence of PSFs and their configurations on RT.

\section{Discussion and Future Work}
\label{discussion}
\atosensedata~provides both the raw and post-processed versions of the data from both the studies. The non-zero probability of not reacting to a stimuli is an important statistic to be recorded and should be documented in real-time through a sensitivity study for future experiments. This could aid in an extensive probabilistic study. To the best of the authors' knowledge, \atosensedata~is the first open dataset of simulator measurements of human train driver performance measurements. It provides a precedent case for the measurement method of simulation and the application of measurement metrics such as RT. While a comparable baseline in a real-world setting could be beneficial for analysis, to the best of our knowledge, no such studies were found (though a relatively similar study was performed in Japan~\cite{wada2020}). This could be one avenue for future research.

Based on analysis results from \atosensedata, standards complementing CSM-RA can be developed. Especially, the development of DIN SPEC 91516 to a full standard can profit from that dataset, since the results will help to set the targets for reference system human driver. Project-related code is available at https://gitlab.opencode.de/dzsf/dzsf-fachbereich-85/ato/atosense). The certification of software, especially AI, for driverless trains requires clear performance targets which can be further used to define essential requirements for the CV system's development. All in all, datasets like \atosensedata~facilitate progress in the field of open track guided transport automation.

\section{Authors' Contributions}
Rustam Tagiew worked on the initial draft of the paper and Prasannavenkatesh Balaji worked on the final version.
\bibliographystyle{ieeetr}
\bibliography{main}
\end{document}